\documentclass[conference]{IEEEtran}
\IEEEoverridecommandlockouts
\usepackage{nicefrac}
\usepackage{caption}
\usepackage{cite}
\usepackage{amsmath,amssymb,amsfonts}
\usepackage{algorithmic}
\usepackage{graphicx}
\usepackage{textcomp}
\usepackage{xcolor}
\usepackage{soul}
\usepackage{xspace}
\usepackage{subfigure}
\newcommand*\rfrac[2]{{}^{#1}\!/_{#2}}
\usepackage{booktabs}
\usepackage{threeparttable}
\def\BibTeX{{\rm B\kern-.05em{\sc i\kern-.025em b}\kern-.08em
    T\kern-.1667em\lower.7ex\hbox{E}\kern-.125emX}}

\definecolor{stelios_colour}{RGB}{200, 238, 200}
\definecolor{light_red}{RGB}{255, 204, 204}

\definecolor{crimson}{rgb}{0.86, 0.08, 0.24}

\newcommand{\tool}{MultiTASC\xspace}

\newif\ifcomment

\commenttrue
\ifcomment
\newcommand{\stelios}[1]{\sethlcolor{stelios_colour}\hl{[\textbf{Stelios:} #1]}}
\newcommand{\sokratis}[1]{\sethlcolor{orange}\hl{[Sokratis: #1]}}

\else
\newcommand{\stelios}[1]{}
\newcommand{\sokratis}[1]{}

\fi

\newcommand{\Hquad}{\hspace{0.5em}} 
    
\begin{document}
% \title{MultiTASC: A Multi-Tenancy-Aware Scheduler for Collaborative DNN Inference}

\title{MultiTASC: A Multi-Tenancy-Aware Scheduler for Cascaded DNN Inference at the Consumer Edge}

\author{Sokratis~Nikolaidis\IEEEauthorrefmark{2},
        Stylianos~I.~Venieris\IEEEauthorrefmark{3},
        and~Iakovos~S.~Venieris\IEEEauthorrefmark{2}%
\\
% sokratisnikolaidis@mail.ntua.gr, s.venieris@samsung.com, venieris@cs.ece.ntua.gr
% \thanks{\IEEEauthorrefmark{2}National Technical University of Athens, Athens, Greece}%
% \thanks{\IEEEauthorrefmark{3}Samsung AI Center, Cambridge, UK}%

\IEEEauthorblockA{\IEEEauthorrefmark{2}National Technical University of Athens, Athens, Greece, \IEEEauthorrefmark{3}Samsung AI Center, Cambridge, UK
}
\IEEEauthorblockA{Email: sokratisnikolaidis@mail.ntua.gr, s.venieris@samsung.com, venieris@cs.ece.ntua.gr}
\vspace{-0.8cm}
}
% \and
% \and

\maketitle

\begin{abstract}
    % Radical progress in the field of on-device execution of Deep Learning (DL) inference tasks has pushed DL deployment from cloud to mobile devices. Despite the advancements, many low-end devices still struggle to execute DL tasks like in the case of indoor intelligence environments. An edge-based server collaborative system can be successful in enabling low-end devices to achieve state-of-the-art accuracy without the need of having the resources of flagship models. It is important to consider the workload that such a scheme would put onto the server when a multitude of devices are being assisted simultaneously. \tool implements a scheduler for a Cascade collaborative architecture, that adapts the forwarding decision functions of devices in real-time to adjust the flow of requests to the server and ensure its responsiveness and full utilization by trading off accuracy when needed.
% 
Cascade systems comprise a two-model sequence, with a lightweight model processing all samples and a heavier, higher-accuracy model conditionally refining harder samples to improve accuracy. By placing the light model on the device side and the heavy model on a server, model cascades constitute a widely used distributed inference approach. 
% Existing cascade methods have focused on single-device cascade where a single device uses the server. 
%With the rapid expansion of intelligent indoor environments, such as smart
%homes, the new setting of Multi-Device Cascade is emerging where multiple and diverse devices are simultaneously using a shared heavy model on the same server, typically located close to the consumer environment. This work presents \tool, a multi-tenancy-aware scheduler that adaptively controls the forwarding decision functions of the devices in order to maximize the system throughput, while sustaining high accuracy and low latency. By explicitly considering device heterogeneity, our scheduler improves the latency service-level objective (SLO) satisfaction rate over state-of-the-art cascade methods in highly heterogeneous setups, while serving over 40 devices, showcasing its scalability.
With the rapid expansion of intelligent indoor environments, such as smart homes, the new setting of Multi-Device Cascade is emerging where multiple and diverse devices are to simultaneously use a shared heavy model on the same server, typically located within or close to the consumer environment. This work presents \tool, a multi-tenancy-aware scheduler that adaptively controls the forwarding decision functions of the devices in order to maximize the system throughput, while sustaining high accuracy and low latency. By explicitly considering device heterogeneity, our scheduler improves the latency service-level objective (SLO) satisfaction rate by 20-25 percentage points (pp) over state-of-the-art cascade methods in highly heterogeneous setups, while serving over 40 devices, showcasing its scalability.
\end{abstract}

% \begin{IEEEkeywords}
% \end{IEEEkeywords}

\vspace{-0.4cm}
\section{Introduction}
\label{sec:intro}
In recent years, there has been significant progress in the field of on-device execution of deep learning (DL) inference tasks~\cite{oodin2021smartcomp}. At the same time, with the rapid expansion of indoor intelligent environments~\cite{laskaridis2022future}, such as smart homes and offices, DL is poised to enable new use-cases by expanding to a greater variety of smart devices, such as IoT cameras and AI speakers. Nevertheless, due to their form-factor and energy-efficiency constraints, most of these devices lie on the low end of the computational spectrum. In contrast to modern high-end smartphones, which host powerful processors and accelerators (GPUs, NPUs)~\cite{smart2021imc}, low-end devices are not able to deploy state-of-the-art deep neural networks (DNNs), resorting to lightweight, but lower-accuracy models.

Given that offloading data to the cloud for inference can incur significant costs in terms of bandwidth, latency and privacy, an alternative scheme is emerging that places the server inside or closer to the consumer environment in the form of a dedicated AI hub that assists the surrounding devices~\cite{laskaridis2022future}. In this context, a prominent deployment approach are the cascade architectures~\cite{park2015big,li2021appealnet,wang2017idk,mirzadeh2020optimal,stamoulis2018designing,kouris2018cascade}. Cascade architectures make use of the fact that not all samples are of the same difficulty and choose to process only the more challenging cases with a powerful model deployed on the server, while letting easier samples, which usually form the majority of the data, to be processed on-device with a light model. A lot of research has been conducted on such architectures, mainly focusing on the forwarding decision criterion and selection of model pairs, progressing the potency of the scheme. 

Despite the progress, the majority of these works have solely focused on the setting where the server is used by a \textit{single} device at any given time. Such an assumption no longer holds in upcoming intelligent environments where multiple devices execute DL inference tasks simultaneously under the support of the same AI hub~\cite{NaShFuKaSh2022}. This gives rise to the new setting of \textit{Multi-Device Cascade}, where multiple devices use the same model on a shared edge-based server. Such a system needs to be scalable in terms of number of devices, balancing fast response time and high accuracy across them. In this context, status-quo approaches, which treat each model cascade independently, would either lead to brute-forcing inference requests through the server's resources, resulting in the system being overwhelmed, or force all devices to fallback to on-device execution, negating any accuracy benefits. Therefore, there is an emerging need for novel methods that explicitly target the challenges of a Multi-Device Cascade.

In this work, we propose \tool, a multi-tenancy-aware scheduler that allows the Multi-Device Cascade architecture to adapt to dynamic conditions by reconfiguring on-the-fly the forwarding criterion of the cascades, thus controlling the server's inference request rate at run time. To sustain smooth operation under device heterogeneity, we further introduce a heterogeneity-aware prioritization scheme that selectively adapts each device's operation depending on its capabilities. Overall, \tool sustains high responsiveness, throughput and accuracy while the number of assisted devices scales in both homogeneous and heterogeneous device ecosystems. The key contributions of this paper are the following:
\begin{itemize}
    \item A system model of the Multi-Device Cascade architecture. By expanding the cascade architecture to accommodate multiple devices, our parametrization exposes the tunable parameters and enables system designers to systematically investigate its trade-offs.
    \item A multi-tenancy-aware scheduler optimized for the Multi-Device Cascade architecture. The proposed scheduler aims to maximize throughput and accuracy while satisfying a latency constraint. This is accomplished through the adaptive manipulation of the forwarding decision functions of the devices that control the inference request flow to the server. By introducing a new metric, Capacity, our scheduler estimates the maximum amount of samples that can be processed on the server within a given latency constraint and utilizes it to dynamically reconfigure the forwarding decision functions on the assisted devices. 
\end{itemize}

\section{Background \& Related Work}
\label{sec:related_work}

\noindent
\textbf{On-Device DNN Inference.}
In recent years, the on-device deployment of DL models has rapidly gained ground~\cite{oodin2021smartcomp}. Although on-device training remains a distant possibility, inference has been achieved on computationally constrained devices through the utilization of various methods, such as lightweight model design~\cite{sandler2018mobilenetv2}, quantization~\cite{han2015deep}, pruning~\cite{he2017channel}, knowledge distillation~\cite{hinton2015distilling}, and optimized scheduling~\cite{CoBiCh2022}. Still, despite the maturity of these techniques, upcoming intelligent spaces, such as smart homes and offices, are often populated with small-form factor, resource-constrained devices (\textit{e.g.}~smart cameras, AI speakers), which lack the processing power to support high-accuracy, computationally intensive models. This fact has motivated the development of distributed collaborative inference approaches. 

\noindent
\textbf{Distributed Collaborative Inference.}
Distributed inference systems employ a server to assist mobile and embedded devices in performing DNN inference tasks. This approach has given rise to two main schemes: \textit{offloading} and \textit{cascading}. \mbox{Offloading}~\cite{almeida2022dyno, huang2020clio, kang2017neurosurgeon} leverages the server's resources to alleviate part of the computational load on the device by splitting the DNN model into two parts; the first part is executed on the device, with the intermediate results forwarded to the server to proceed with the execution of the second part. 

In the cascade scheme, samples are fed into a two-model sequence of DNNs, with progressively increasing complexity and accuracy. Once the input is processed by a model, its output is evaluated by a forwarding decision function, which determines whether the inference ends using the current result or proceeds to the next model. Several works have been conducted on this area. \cite{park2015big} proposes to analyze the difference between the best and the second best softmax result in order to determine which inputs can be processed only by the light model and which require the heavy model. \cite{li2021appealnet} introduces a trainable forwarding criterion by training a neural head on the light model's feature extractor. \cite{wang2017idk} explores the idea of using more than two DNN models and compares different decision metrics. Finally, \cite{mirzadeh2020optimal} and \cite{stamoulis2018designing} propose solutions for deploying cascades under tight energy constraints. 

\noindent
\textbf{Multi-Device Cascades.}
Previous works on cascades~\cite{park2015big,li2021appealnet,wang2017idk,mirzadeh2020optimal,stamoulis2018designing} have focused on an isolated setting, where a single device has exclusive access to a server. Nonetheless, with an increased rate of AI-enabled applications deployed in indoor environments, this assumption is unrealistic; instead, multiple devices will require support from the same server at any given time. Such a setting requires principled investigation, since a brute-force deployment, where devices run independently of each other, would lead to either an overloaded server, and in turn to long response times, or significant drop in accuracy by falling back to local execution. The Multi-Device Cascade setting remains unexplored and is the focus of this work. 

\section{Multi-Device Cascade of Classifiers}
\label{sec:system_model}
\begin{figure}[t]
    \vspace{0.1cm}
    \centerline{\includegraphics[scale=0.7,trim=6cm 0cm 6cm 0cm]{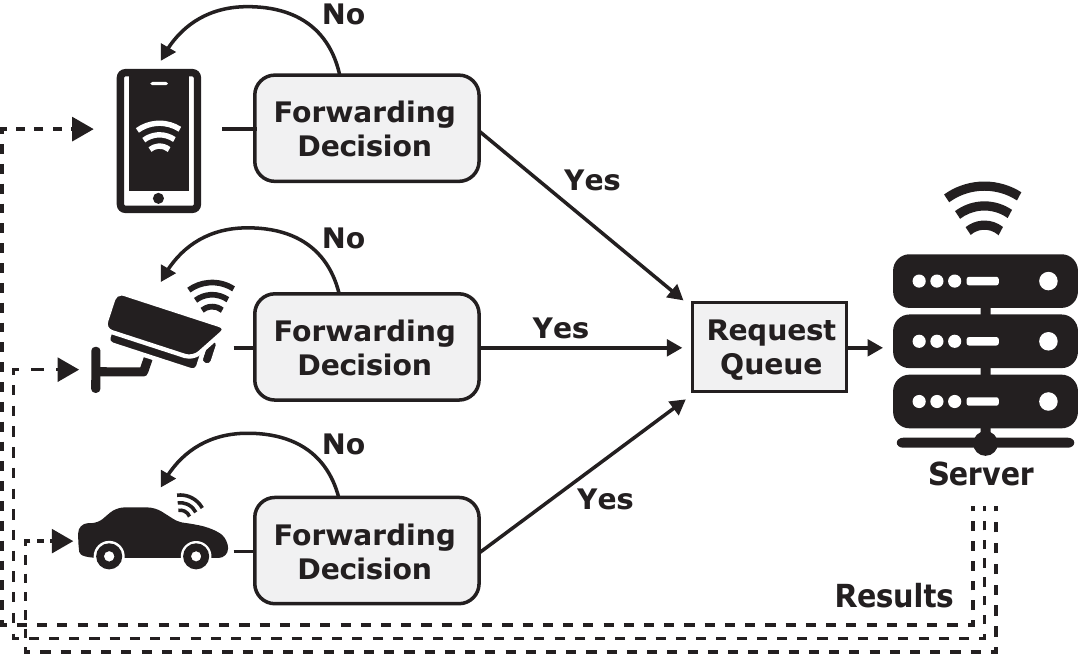}}
    \caption{\small System architecture of Multi-Device Cascade.}
    \label{fig:sys_arch}
    \vspace{-0.4cm}
\end{figure}

Fig.~\ref{fig:sys_arch} presents the system architecture of a Multi-Device Cascade, where IoT devices are running DL inference tasks. All devices perform the same task, \textit{e.g.}~object detection, but may host different models. The output predictions of each sample are passed on to a \textit{Forwarding Decision} function that determines whether the DL model of the device is confident about its output. Depending on the decision, either the result remains as is, or the sample is forwarded to a server to be processed by a more accurate model. The forwarded samples from all devices are put into a \textit{Request Queue} from where they are drawn to form the input of the server-hosted model; as such, the server-side model is \textit{shared} among all devices. Finally, the results produced on the server are sent back to their corresponding devices.

\noindent
\textbf{Single-Device Cascade.} Let $x$$\in$$\mathcal{X}$ be the input of a DL inference task performed on an IoT device and $y$$\in$$\{1, ..., K \}$ the classification label produced by the model, where $K$ is the number of classes. By using a decision function, $d(\cdot)$,  on the output of the light device model, we can decide whether the result is satisfactory ($d(\cdot)$$=$$0$) or we should forward the sample to the heavier model for further processing ($d(\cdot)$$=$$1$). Denoting the classification function of the light model by \mbox{$f_l: \mathcal{X} \rightarrow [0,1]^{K}$} that yields the softmax ouput vector of the model whose maximum value is the predicted class, and the classification of the heavy model by \mbox{$f_h: \mathcal{X} \rightarrow [0,1]^K$}, we formally define a collaborative cascade system as: 
\begin{equation}
\it{casc}_{f_l, f_h, d} (x) =  
 \begin{cases} 
      f_l(x) &\text{if} \quad \, d(f_l(x)) = 0 \\
      f_h(x) &\text{if} \quad \, d(f_l(x)) = 1 
\end{cases}
\label{cascade}
\end{equation}
\textbf{Multi-Device Cascade.} To capture Multi-Device Cascade architectures (Fig.~\ref{fig:sys_arch}), we extend the single-device cascade system modeling as follows. Let $\mathcal{D}$ be the set of devices that use the server as a collaborator. Then, the Multi-Device Cascade system is defined as:
\begin{equation*}
    \small
    \it{casc}_{f_l^i, f_h, d^i} (x^i) =  
     \begin{cases} 
          f_l^i(x^i) &\text{if} \Hquad \, d^i(f_l^i(x)) = 0 \Hquad \forall i \in \{1, ..., |\mathcal{D}|\} \\
          f_h(x^i) &\text{if} \Hquad \, d^i(f_l^i(x)) = 1 %\quad \{1, ..., |\mathcal{D}|\}
    \end{cases}
    \label{multicascade}
\end{equation*}
where $x^i$$\in$$\mathcal{X}^i$ is a sample processed by the i-th device, $f_l^i$ the classification function of the DL model deployed on the i-th device, $f_h$ the shared heavy model on the server, and $d^i(f_l^i(x))$ the forwarding decision function of the i-th device.

\noindent
\textbf{Congestion Problem.} When a single device is using the server as the collaborator, it has exclusive access to the server's computational resources and hence the response time is minimized. With the proliferation of IoT devices, such a server should be utilized by many devices at the same time, to amortize its cost and provide maximum utility. Nonetheless, depending on the conditions, if the arrival rate of incoming requests exceeds the attainable processing throughput of the server, the server will be overloaded and the requests will experience severe waiting times in the request queue.

Given the number of devices $|\mathcal{D}|$, we express the arrival rate of requests to the server as $AR_{\text{server}} = \sum_{i~=~1}^{|\mathcal{D}|} \rfrac{p_{\text{casc}}^i}{t_{\text{inf}}^i}$
%
% \begin{equation}
%     AR_{\text{server}} = \sum_{i~=~1}^{|\mathcal{D}|} \frac{p_{\text{casc}}^i}{t_{\text{inf}}^i}
%     \label{ar}
% \end{equation}
% 
where $t_{\text{inf}}^i$ is the average inference latency of a sample on the i-th device and $p_{\text{casc}}^i$ is the probability of a sample giving $d^i(f_l^i(x^i))=1$. 
Given the attainable throughput $T_{\text{server}}$ of the server, we distinguish between three different states: 
\begin{itemize}
    \item $AR_{\text{server}} < T_{\text{server}}$: the server's processing rate is larger than the arrival rate, resulting in the server being underutilized. A larger number of difficult samples could be sent to the server to achieve higher accuracy. 
    \item $AR_{\text{server}} = T_{\text{server}}$: equilibrium is attained. Requests are processed upon arrival without accumulating and the server's processing power is fully utilized. 
    \item $AR_{\text{server}} > T_{\text{server}}$: the requests arrive faster than the server can process. If this state lasts, the request queue will be large, resulting in unwanted latency.
\end{itemize}
Since the probability $p_{\text{casc}}^i$ of the forwarding decision function is not static, but changes during inference, such an architecture could benefit by dynamically adapting its state depending on the current conditions. Since $t_{\text{inf}}^i$ and $T_{\text{server}}$ are fixed based on the device and server-side processors, we opt to manipulate $p_{\text{casc}}^i$ by changing the parameters of $d^i(f_l^i(x^i))$ in order to introduce adaptability to the system.

\noindent
\textbf{Problem Optimization.} We formulate the aforementioned problem as a multi-objective optimization problem, aiming to maximize accuracy and throughput subject to a latency service-level objective (SLO). The next section describes our proposed scheduler that tackles this problem.

\section{Proposed solution}
\label{sec:scheduler}
\begin{figure}[tbp]
    \vspace{0.1cm}
    \centerline{\includegraphics[scale=0.9,trim=6cm 0cm 6cm 0cm]{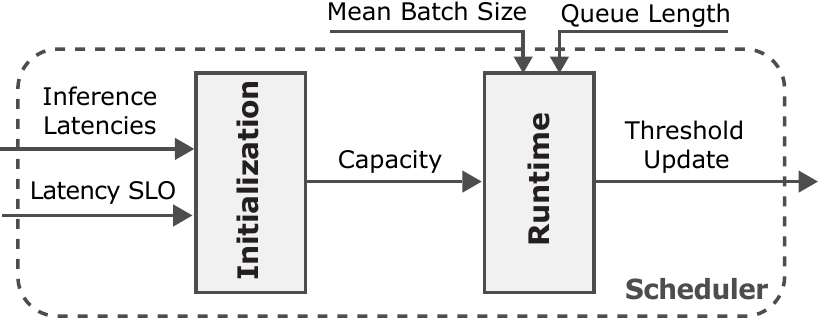}}
    \caption{\small \tool's internal architecture.}
    \label{fig:shed_arch}
    \vspace{-0.4cm}
\end{figure}

To combat the accumulation of requests or underutilization of server resources, we propose \tool, a multi-tenancy-aware scheduler that dynamically adapts the forwarding decision functions on assisted devices in order to control the arrival rate of samples. Fig.~\ref{fig:shed_arch} depicts its internal design. The scheduler monitors the state of the request queue, communicates with the assisted devices and tunes the flow of incoming requests based on the current conditions. To this end, we introduce four techniques: \textit{i)}~reconfigurable forwarding decision functions, \textit{ii)}~the Capacity metric on the server, \textit{iii)}~fractional update, and \textit{iv)}~device heterogeneity-aware prioritization for effectively communicating updates to the devices. 

\subsection{Reconfigurable Forwarding Decision Function}
\label{sec:fwd_dec_func}

Research effort has been invested into quantifying the prediction confidence of DNNs, leading to several approaches~\cite{guo2017calibration, wang2017idk}. In this work, we adopt the Best-versus-Second-Best (BvSB) metric~\cite{bvsb2009cvpr}, using the difference between the two highest values of the softmax outputs of the model 
as
\mbox{$\text{BvSB}\Bigr|_{f(x)}$$=$$P_1 - P_2$},
% 
% \begin{equation}
%     \text{BvSB}\Bigr|_{f(x)} = P_1 - P_2
%     \label{eq:bvsb}
% \end{equation}
%
where $P_1$ and $P_2$ are the maximum and second maximum values, respectively, in the output softmax vector of classifier $f(x)$.
Other metrics, such as top-1 softmax or entropy, can be used with minimal modifications, potentially leading to different latency-accuracy trade-offs. 

The vast majority of existing cascade systems opts to select a specific threshold at design time, which then remains fixed upon deployment. In this work, to accommodate the adaptability needs of our target system, we adopt an alternative scheme where the decision function can be \textit{dynamically reconfigured}. The decision function $d^i(\cdot)$ is thus defined as: %\mbox{$ 0 ~\text{if} ~\text{BvSB}\Bigr|_{f^i_l(x)} \geq c^{i,t}, ~\text{or} ~1 ~\text{if} $} \mbox{$~\text{BvSB}\Bigr|_{f^i_l(x)} < c^{i,t}$}
\begin{equation}
d^i(f^i_l(x)) =  
 \begin{cases} 
      0 &\text{if} \quad \text{BvSB}\Bigr|_{f^i_l(x)} \geq c^{i,t}\\
      1 &\text{if} \quad \text{BvSB}\Bigr|_{f^i_l(x)} < c^{i,t} 
\end{cases}
\label{decisionfunc}
\end{equation}
where $c^{i,t}$ is the decision threshold of device $i$ at time $t$.
The per-device decision thresholds are exposed to our server-residing scheduler, which adapts them at run time.

\subsection{The Capacity Metric}
\label{sec:capacity}

To fully take advantage of the server's computational resources and boost throughput, it is important to use batching, \textit{i.e.}~processing multiple samples at the same time. To avoid the latency that would arise from waiting for the request queue to reach a specific batch size, we employ dynamic batching~\cite{ali2020batch}. With dynamic batching, we use the maximum batch size that is feasible with the current request queue length. Available batch sizes are $\mathcal{B}$$=$$\{1,2,4,8,16,32,64\}$. Due to diminishing returns, in some cases we use a lower maximum batch size, \textit{e.g.}~with EfficientNetB3 a batch size of 16 provides higher throughput and lower latency than a batch size of 32 and above.

To calculate the amount of samples that can be processed by the server within a given latency constraint, we introduce the Capacity metric. Based on the inference latency of each batch size and a given latency SLO, the scheduler calculates the maximum amount of samples that can be classified without latency violations. We call that amount Capacity. 

To obtain the value of the server's Capacity, we cast the problem as 
% Finding the server's Capacity is 
an unbounded variation of the Knapsack problem, where 
the batch size throughput is analogous to the value/weight ratio. Since we know that the larger the batch size the higher the throughput (up to the point where the server's processing power is saturated), we introduce a greedy algorithm that adds the largest batch size as many times as possible. Let $\mathcal{B}$ be the pool of batch sizes, then:
\begin{equation}
   C = \max_{n} \sum_{j=1}^{|\mathcal{B}|} b_j n_j, \label{capacity}  \quad \text{s.t.} \quad \sum_{j=1}^{|\mathcal{B}|} L_{\text{inf}}^{b_j} ~ n_j \leq L^{\text{SLO}}, \, n_j \geq 0
\end{equation}
where $C$ is the Capacity metric, $b_j$ is the j-th batch size, $n_j$ is the amount of times the batch size is used, $L_{\text{inf}}^{b_j}$ is the inference latency of the j-th batch size and $L^{\text{SLO}}$ is the latency SLO.

\begin{table}[t]
\vspace{0.4cm}
    \captionsetup{font=small,labelfont=bf}
    \caption{\normalsize Evaluated DNN Models}
    \centering
    \resizebox{0.475\textwidth}{!}{
    \setlength{\tabcolsep}{2pt}
    \begin{threeparttable}
        \begin{tabular}{lllcllrr}
            \toprule
            \textbf{Model} & \textbf{Location} & \textbf{Device} & \textbf{Clock Rate} & \textbf{Accuracy} & \textbf{Latency} & \textbf{FLOPs} & \textbf{\#Params}\\
            \midrule
            MobileNetV2 & Low-end & Sony Xperia C5 &  1.69 GHz & 71.85\% & 31 ms & 0.6 B & 3.5 M \\
            EfficientNetLite0 & Mid-tier & Samsung A71 & 2.20 GHz & 75.02\% & 43 ms & 0.8 B & 4.7 M \\
            EfficientNetB0 & High-end & Samsung S20 FE &  2.73 GHz & 77.04\% & 33 ms & 0.8 B & 5.3 M \\
            InceptionV3 & Server & Tesla T4 GPU & 585 MHz & 78.29\% & 15 ms & 11.4 B & 23.8 M \\
            EfficientNetB3 & Server & Tesla T4 GPU & 585 MHz & 81.49\% & 25 ms & 3.7 B & 12.2 M \\
            \bottomrule
        \end{tabular}
        \begin{tablenotes}
            \small
            \item * See Table 1 in~\cite{oodin2021smartcomp} for the detailed resource characteristics of the target mobile phones.
        \end{tablenotes}
    \end{threeparttable}
    }
    \label{modeltable}

\vspace{-0.4cm}
\end{table}

\subsection{Fractional Update}
\label{sec:perc_change}

Changing the thresholds of all devices at once could lead to unwanted, sudden oscillations of the system. To achieve a smoother operation of the system and give enough time for each adaptation step to affect the execution, \tool introduces the \textit{fractional update} technique. With fractional update, \tool updates the thresholds of only a certain percentage, denoted by $P$, of the total number of devices for each update. Thresholds are updated as dictated by Eq.~(\ref{percentage}). We set the invocation rate of the scheduler to be once every 2~seconds, allowing for update results to affect the system.
% 
%Depending on whether the threshold is increasing or decreasing, devices with lower or higher threshold, accordingly, are given priority, so that all devices have similar thresholds. 
\begin{equation}
    \it{TC} = 
    \begin{cases}
        -M & \text{if}  \quad \Bar{b} > \alpha \cdot C \quad \text{and} \quad QL > \alpha \cdot C \\
        +M & \text{if} \quad \Bar{b} \leq \beta \cdot C \quad \text{and} \quad QL \leq \beta \cdot C \\
      ~ ~  0 &  \text{otherwise}
    \end{cases}
    \label{percentage}
\end{equation}
where $M$ is the margin by which the threshold of the chosen devices changes, $\Bar{b}$ the average of the last $L$ batch sizes used for inference on the server, $C$ the Capacity metric and $QL$ the request queue length. Capacity is weighted using the parameters $\alpha$ and $\beta$. If requests accumulate beyond a certain limit despite the dynamic threshold updates, all thresholds are set to 0 until the server is decongested.

\subsection{Device Heterogeneity-Aware Prioritization}
\label{sec:types_change}

\tool explicitly considers device heterogeneity in order to maximize both the average accuracy and the system throughput. Our heterogeneity-aware prioritization strategy updates the threshold values using Eq.~(\ref{percentage}), but expands the pipeline by selecting which type of devices receive the update. The key insight of our strategy is that when it comes to decreasing the thresholds, \tool prioritizes high-end and mid-tier devices since they host larger models and can maintain higher accuracy even with low thresholds. In contrast, when increasing thresholds, our scheduler prioritizes low-end devices, since they are the ones benefiting the most from higher thresholds due to hosting lighter models.

% \subsection{Limit}

% To avoid extreme request accumulation, we implemented a hard limit for the length of the request queue. If the limit is reached, all thresholds are set to 0 and \tool allows them to increase only after the queue is decongested. 

\section{Evaluation}
\label{sec:eval}
% \begin{figure}[t]
%     \centerline{\includegraphics[scale=0.4,trim=6cm 0cm 6cm 0cm]{figures/Plots/MobEffSLOFill.pdf}}
%     \caption{SLO satisfaction rate for MobileNetV2-EfficientNetB3 pair.}
%     \label{plot:Mob_Eff_SLO}
%     \vspace{-0.4cm}
% \end{figure}

% \begin{figure}[t]
%     \centerline{\includegraphics[scale=0.4,trim=6cm 0cm 6cm 0cm]{figures/Plots/MobEffTpAccFill.pdf}}
%     \caption{Throughput and accuracy for MobileNetV2-EfficientNetB3 pair.}
%     \label{plot:Mob_Eff_TpAcc}
%     \vspace{-0.4cm}
% \end{figure}

\subsection{Experimental Setup}
\label{sec:exp_setup}
To evaluate the performance of \tool, we built a prototype on top of TensorFlow 2.9.1 targeting an edge server and three tiers of mobile devices. The edge server hosts an NVIDIA Tesla T4 GPU, Intel(R) Xeon(R) 2.30GHz CPU and 12GB of RAM. For the clients, we target three smartphones of increasing processing capabilities, namely: Sony Xperia C5 Ultra,  Samsung A71 and Samsung S20 FE, representing low-, mid- and high-end clients, respectively. To assess our system across various settings, we conduct simulation-based experiments, varying the target latency SLO and the number of client devices. We measure the average inference time across 200 runs on the target devices for the evaluated models. We follow the same approach for the server-side models across different batch sizes. All on-device measurements were performed using TensorFlow Lite and targeting the CPU of the respective mobile device.
For the device-server communication, we employ the AMQP protocol, following the widely used practice for communication between IoT devices.

\noindent
\textbf{Models \& Datasets.}
We target the task of 1k-class image classification. Concretely, we use the ImageNet dataset and its 50k-images validation set in our experiments. Table~\ref{modeltable} shows the evaluated models. We obtain the ImageNet-pretrained models as provided by TensorFlow Hub. On the client side, to emulate the common approach where models are selected based on the capabilities of each device, we deploy MobileNetV2, EfficientNetLite0 and EfficientNetB0 on low-, mid- and high-end devices, respectively. On the server side, we chose InceptionV3 and EfficientNetB3 as representative models that are computationally heavy, but provide high accuracy. 

\begin{figure}[t]
    \centerline{\includegraphics[scale=0.4,trim=6cm 0cm 6cm 0cm]{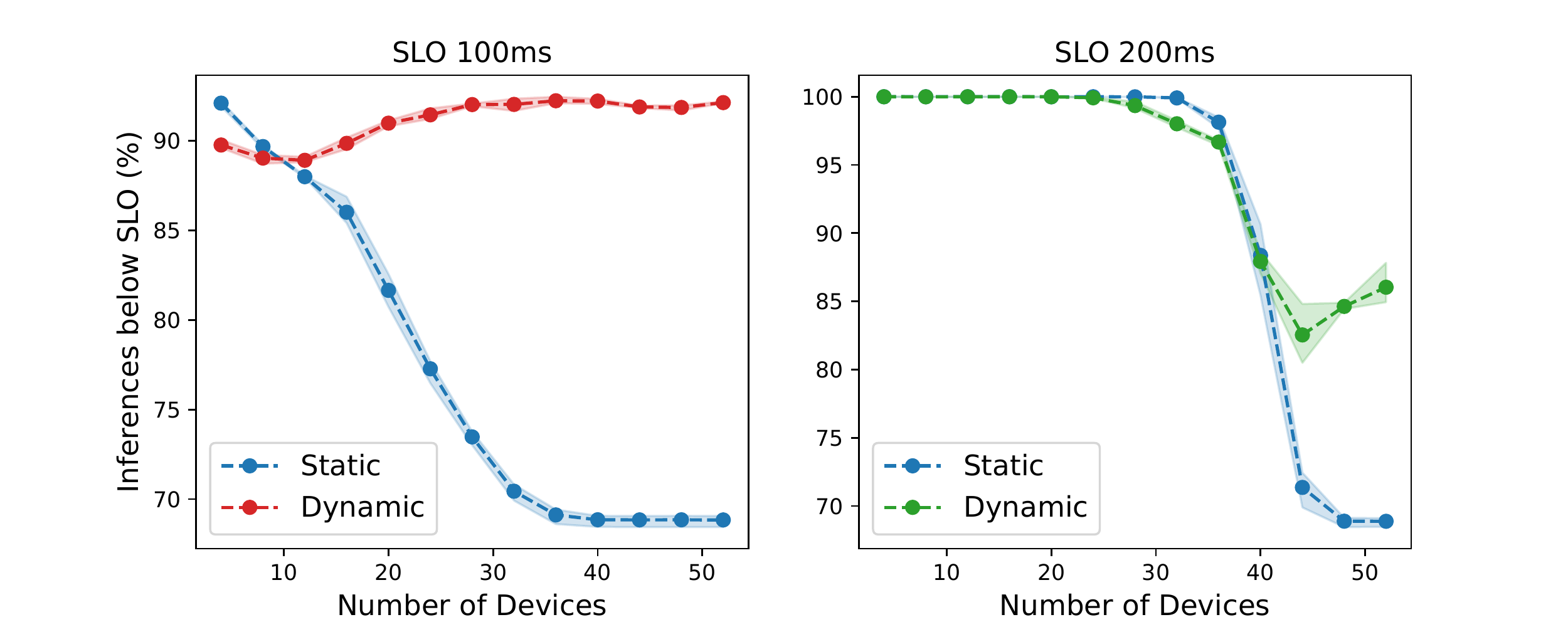}}
    \caption{\small SLO satisfaction rate for EfficientNetLite0-InceptionV3 pair.}
    \label{plot:Eff_Inc_SLO}
    \vspace{-0.4cm}
\end{figure}

\begin{figure}[t]
    \centerline{\includegraphics[scale=0.4,trim=6cm 0cm 6cm 0cm]{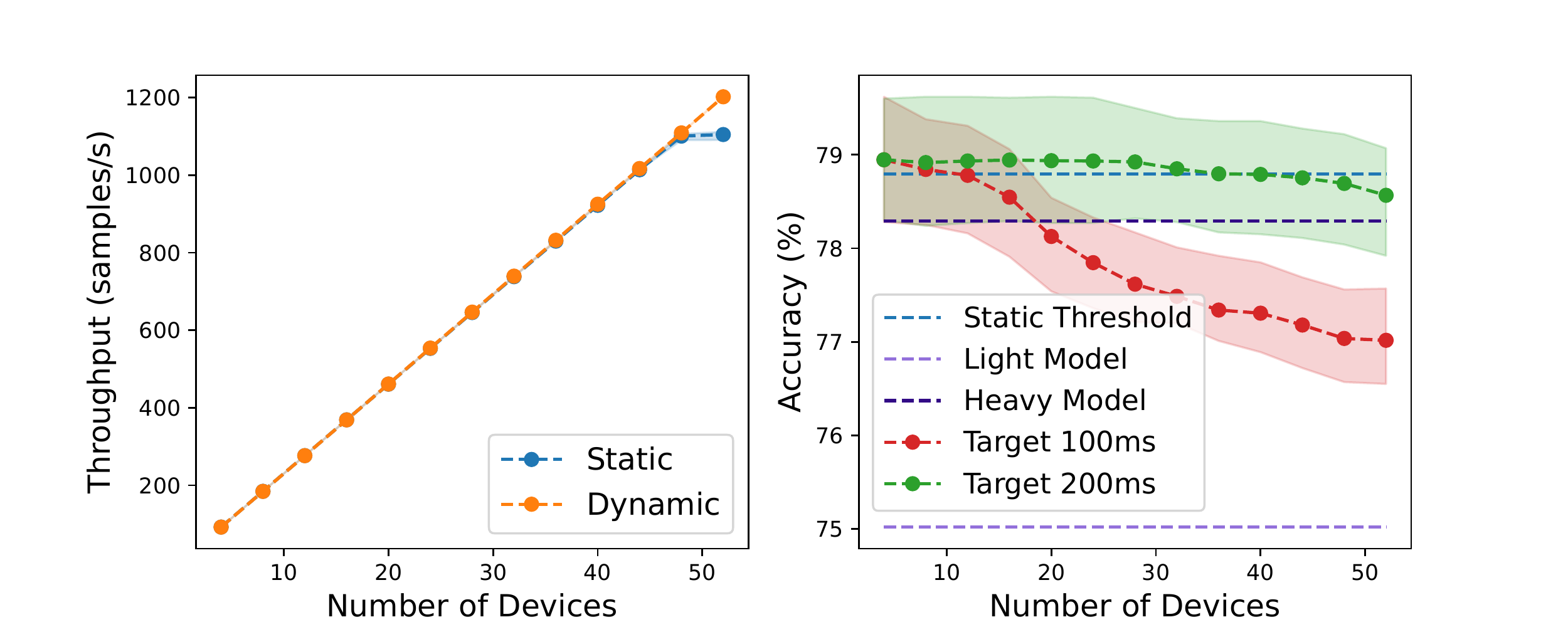}}
    \caption{\small Throughput and accuracy for EfficientNetLite0-InceptionV3 pair.}
    \label{plot:Eff_Inc_TpAcc}
    \vspace{-0.4cm}
\end{figure}

\noindent
\textbf{Evaluation Settings.}
We focus on two types of device ecosystems: \textit{i)}~homogeneous, which comprises devices of equal processing capabilities that host the \textit{same} local model; and \textit{ii)}~heterogeneous, which comprises devices of diverse processing capabilities, with each device hosting a model sized to its tier.
In the homogeneous scenario, all devices were mid-tier, \textit{i.e.}~A71 phones, and ran EfficientNetLite0 with an average on-device inference latency of 43~ms. In the heterogeneous scenario, all three tiers of devices were deployed in equal percentage. In both cases, the dataset of each device consisted of 5,000 randomly selected samples from the last 40,000 images of ImageNet's validation set. Three different seeds were used and the average is reported. 
The metrics used for the evaluation are: the system throughput, the average accuracy across devices, latency SLO satisfaction rate for 100- and 200-ms SLOs,  and scalability in terms of number of devices.

% \textbf{Homogeneous Device Ecosystem.} In the homogeneous scenario, all devices were mid-tier, \textit{i.e.}~A71 phones, and ran either MobileNetV2 or EfficientNetLite0 with an average on-device inference latency of 20~ms and 43~ms, respectively. The dataset of each device, consisted of 5,000 randomly selected samples from the 40,000 last images of ImageNet's validation set. Three different seeds were used and the average is reported. 

% \textbf{Heterogeneous Device Ecosystem.} In the heterogeneous scenario, all three tiers of devices were deployed in equal percentage. As before, 5,000 randomly selected samples from the 40,000 last images of ImageNet's validation set were used as the dataset on each device and the average was reported across three different seeds.

\noindent
\textbf{Baseline.} As a baseline, we use a scheduler with statically selected thresholds that remain fixed at run time. To choose the static threshold, we use the first 10,000 images of ImageNet's validation set as our calibration set and evaluate all cascade model pairs in terms of accuracy and forwarding probability. As such, we tune the threshold so that approximately 30\% of samples are forwarded to the heavy model, providing a balanced accuracy-latency trade-off. In cases where that threshold yielded an accuracy loss of more than 1~pp compared to the highest achievable cascade accuracy, we used the lowest threshold that satisfied the 1~pp limit. This baseline is equivalent to a set of state-of-the-art cascades~\cite{li2021appealnet,wang2017idk,kouris2018cascade}.

\begin{figure}[t]
    % \centerline{\includegraphics[scale=0.28,trim=6cm 0cm 6cm 0cm]{figures/Plots/DiffIncSLO.pdf}}
    \centerline{\includegraphics[scale=0.4,trim=6cm 0cm 6cm 0cm]{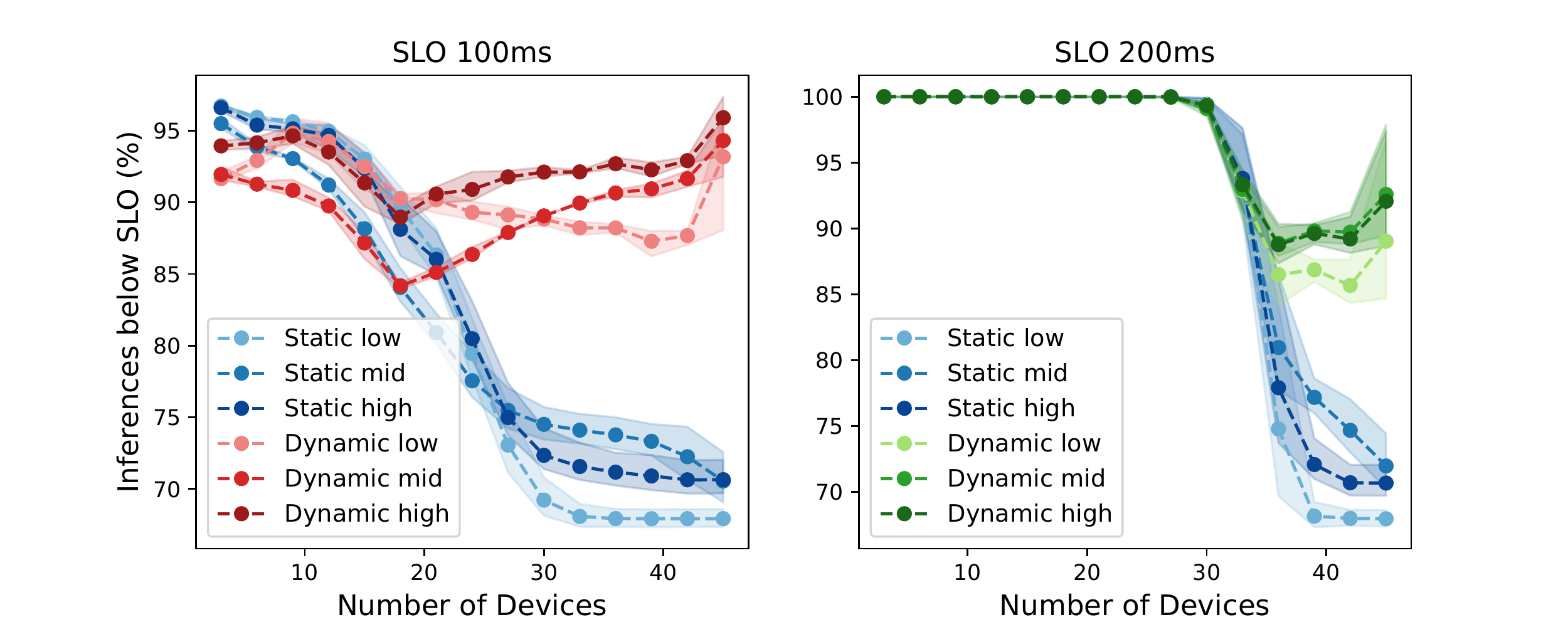}}
    \caption{\small SLO satisfaction rate for InceptionV3 on the server.}
    \label{plot:Diff_Inc_SLO}
    \vspace{-0.4cm}
\end{figure}

\begin{figure}
    \centering
    \subfigure{
        \includegraphics[scale=0.28,trim={6cm 0cm 6cm 0cm},width=0.4\textwidth]{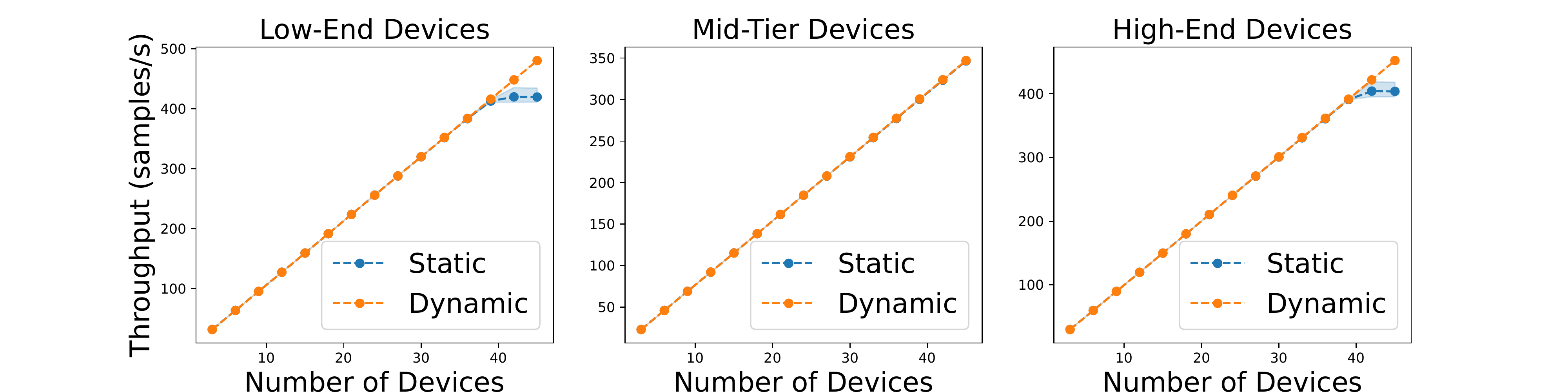}
        \label{plot:Diff_Inc_TP}
    }
    \put (-220,-8) {\footnotesize (a)}
    
    \subfigure{
        \includegraphics[scale=0.28,trim={6cm 0cm 6cm 0cm},width=0.4\textwidth]{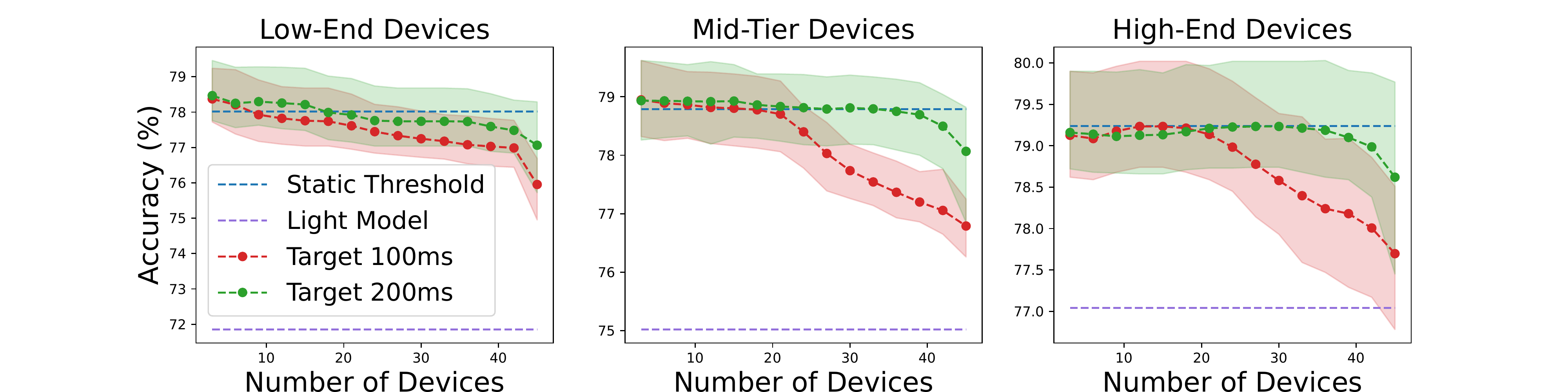}
        \label{plot:Diff_Inc_Acc}
    }
    \put (-220,-8) {\footnotesize (b)}
    \caption{\footnotesize (a) Throughput and (b) accuracy for InceptionV3 on the server.}
    \label{fig:Diff_Inc}
    \vspace{-0.4cm}
\end{figure}

\subsection{Evaluation of \tool's Performance}
\label{sec:eval_perf}

Here, we assess the performance of our scheduler across both device ecosystems. We optimized \tool's parameters $P$, $M$, $L$, $\alpha$ and $\beta$ using grid search and set them equal to 20\%, 0.05, 5, 0.83 and 0.125, respectively.

\begin{figure}[t]
    \centerline{\includegraphics[scale=0.4,trim=6cm 0cm 6cm 0cm]{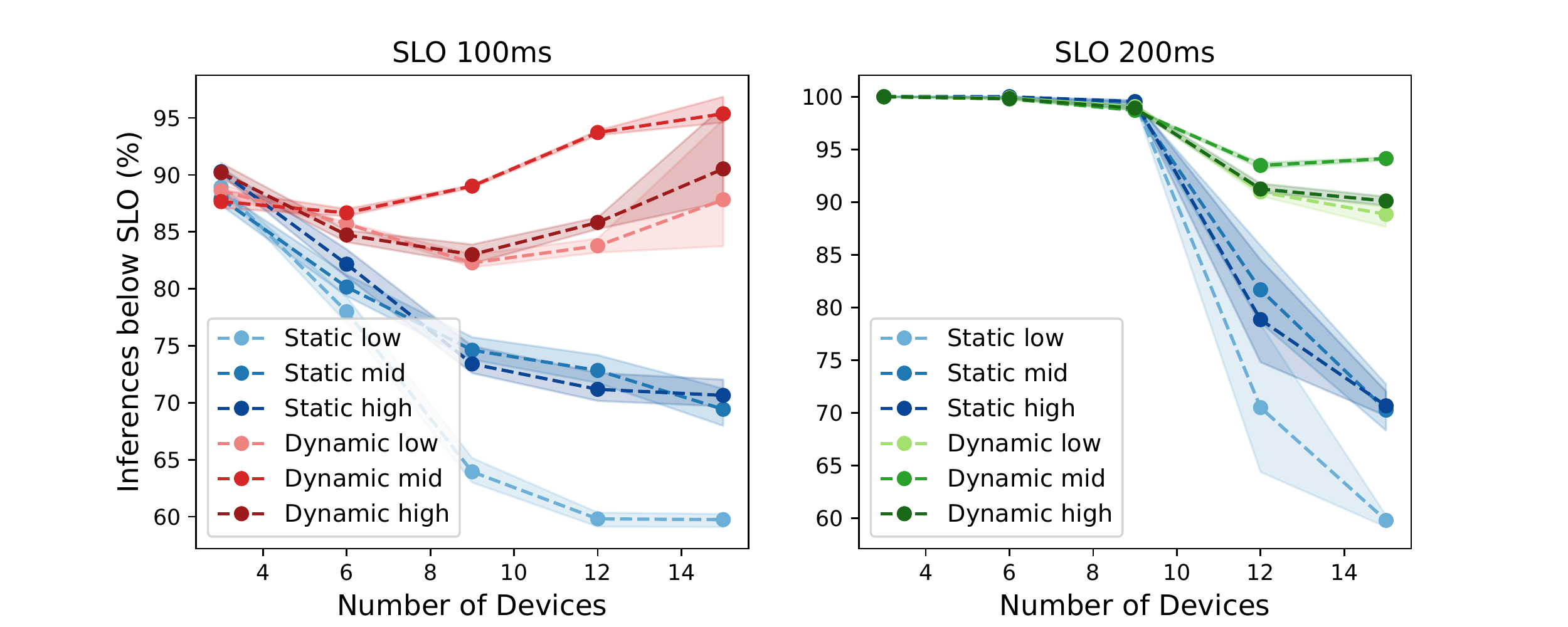}}
    \caption{\small SLO satisfaction rate for EfficientNetB3.}
    \label{plot:Diff_Eff_SLO}
    \vspace{-0.4cm}
\end{figure}

\begin{figure}
    \centering
    \subfigure{
        \includegraphics[scale=0.28,trim={6cm 0cm 6cm 0cm},width=0.4\textwidth]{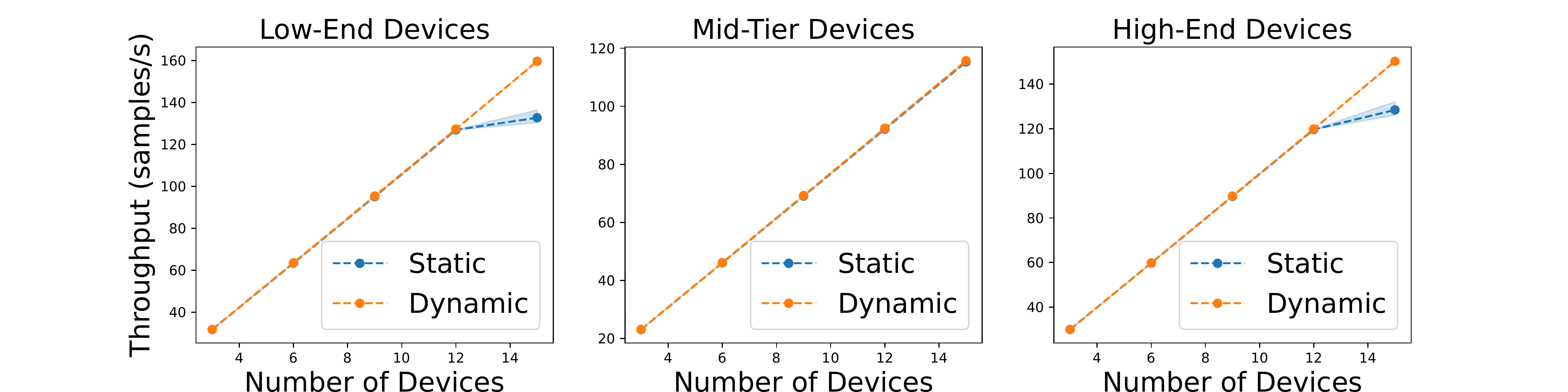}
        \label{plot:Diff_Eff_TP}
    }
    \put (-220,-8) {\footnotesize (a)}
    
    \subfigure{
        \includegraphics[scale=0.28,trim={6cm 0cm 6cm 0cm},width=0.4\textwidth]{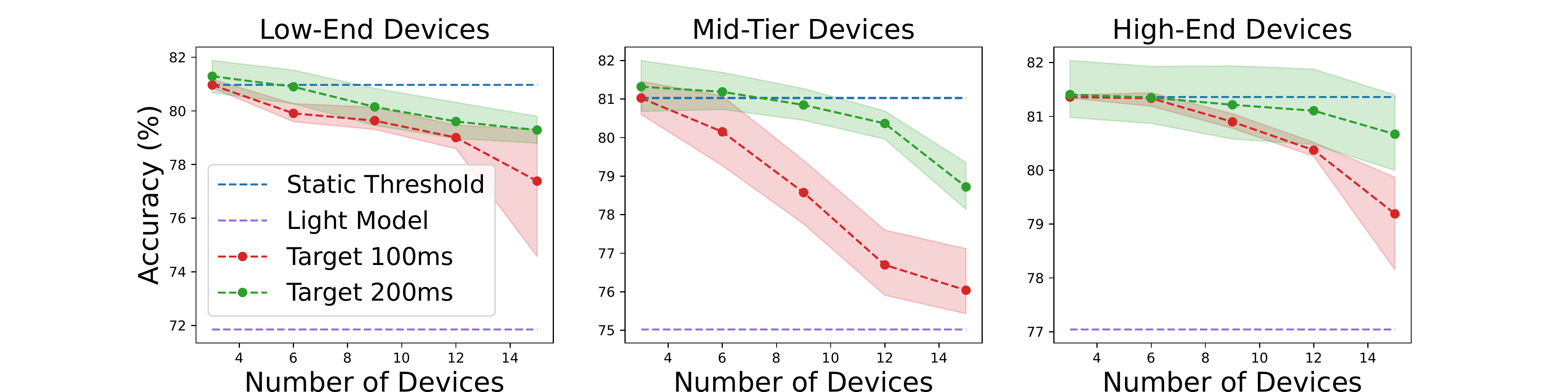}
        \label{plot:Diff_Eff_Acc}
    }
    \put (-220,-8) {\footnotesize (b)}
    \caption{\footnotesize (a) Throughput and (b) accuracy for EfficientNetB3 on the server.}
    \label{fig:Diff_Eff}
    \vspace{-0.4cm}
\end{figure}

\noindent
\textbf{Homogeneous Scenario.} Fig.~\ref{plot:Eff_Inc_SLO} shows the SLO satisfaction rate comparison for the pair EfficientNetLite0-InceptionV3. For a small amount of devices, \tool keeps the satisfaction rate close to the baseline. As the baseline starts failing at 20 and 40 devices for SLOs of 100 and 200~ms, respectively, \tool gradually reconfigures the thresholds of the forwarding decision functions and manages to keep satisfaction rate high, gaining 15-20~pp across the SLOs. In Fig.~\ref{plot:Eff_Inc_TpAcc}, we can see that after 45 devices the baseline's throughput is starting to plateau, due to the queue not being served quickly enough. The users of the devices would experience this plateau as excessive response time. In contrast, with \tool, the aggregate throughput continues to increase linearly with a growing number of devices, indicating that the requests are not allowed to accumulate and more devices can be served. 

With respect to accuracy, for a small number of devices, \tool slightly increases accuracy. This happens because our approach recognizes that the server is being underutilized and increases the thresholds accordingly. As the number of devices becomes larger, accuracy is traded off to achieve lower latency and sustain higher throughput, but still stays within 2~pp of the baseline. Additionally, we observe that the accuracy achieved by the cascade is in most cases higher than the accuracy of the heavy model, showcasing the benefits of classifier cascades for the specific model pair. Moreover, we observe the scalability of \tool, serving up to 50 devices without significant loss of accuracy. Different model pairings gave similar results, providing high satisfaction rate, sustaining the throughput and keeping accuracy within acceptable limits. When comparing server-side models, InceptionV3 can serve a larger amount of devices compared to EfficientNetB3, due to its lower computational demands, but EfficientNetB3 achieves higher accuracy when it comes to a lower number of devices.

We further observe that when the on-device inference latency is below the SLO, the lowest limit for the SLO satisfaction rate is the percentage of samples that are not forwarded to the server. For the EfficientNetLite0-InceptionV3 pair, this is around 69\% and can be observed when the baseline solution reaches 40 devices for the 100-ms SLO (Fig.~\ref{plot:Eff_Inc_SLO})

% Overall, from this set of experiments, we speculate that selecting the highest-performing pair of models depends on the number of devices to be assisted. Specifically, we observe that when a small number of devices are present, an extremely light model paired with an extremely complex model, \textit{e.g.}~MobileNetV2-EfficientNetB3, achieves higher overall accuracy with low response times. On the other hand, when a larger number of devices are assisted, coupling a more complex and accurate device-side model with a mid-sized, higher-throughput model on the server-side, \textit{e.g.}~EfficientNetLite0-InceptionV3, manages to sustain high throughput and low response times with marginal loss of accuracy.  

\noindent
\textbf{Heterogeneous Scenario.} Fig.~\ref{plot:Diff_Inc_SLO}-\ref{fig:Diff_Inc} show the SLO satisfaction rate, throughput and accuracy comparison between \tool and the baseline, when targeting a heterogeneous device environment with InceptionV3 as the server-side model. We report the performance results separately for each device tier. When calculating the throughput, different latency targets give similar results, so we present only one. Similarly to the homogeneous case, we observe that the satisfaction rates are maintained high across all tiers of devices, while the baseline leads to catastrophically degraded latency after 30 and 40~devices for SLOs of 100 and 200~ms, respectively (Fig.~\ref{plot:Diff_Inc_SLO}). Our approach achieves gains of approximately 20-25~pp in satisfaction rate across all device tiers. We also observe the throughput plateau, but only for the low- and high-end devices. This is attributed to the fact that the latency of EfficientNetLite0 on the mid-tier devices is 1.32$\times$ slower than the other tiers, resulting in mid-tier devices continuing inference after the inference on low-end and high-end devices has finished. This allows the server to decongest since only $\nicefrac{1}{3}$ of the devices remains. Accuracy is again within acceptable values, never dropping below 2~pp from the baseline. 

Fig.~\ref{plot:Diff_Eff_SLO}-\ref{fig:Diff_Eff} show the performance comparison when EfficientNetB3 is used as the server-side model. Our observations are similar to the previous case of InceptionV3, showcasing the generality of our approach. From this set of experiments, we infer that the highest-performing server-side model is in most cases the higher-throughput one, \textit{e.g.}~InceptionV3. Specifically, we observe that InceptionV3 can serve a much larger number of devices while keeping accuracy within acceptable levels compared to the complex, low-throughput EfficientNetB3. EfficientNetB3 achieves higher accuracy when the number of devices is 9 or lower for SLO of 100~ms or 14 and lower for SLO of 200~ms. In all cases, via adjusting the forwarding decision thresholds during execution, \tool manages to keep response times low and system throughput high by gradually trading off accuracy.

% \begin{figure}[t]
%     \centerline{\includegraphics[scale=0.28,trim=6cm 0cm 6cm 0cm]{figures/Plots/DiffEffTP.pdf}}
%     \caption{Throughput for EfficientNetB3.}
%     \label{plot:Diff_Eff_TP}
% \end{figure}

% \begin{figure}[t]
%     \centerline{\includegraphics[scale=0.28,trim=6cm 0cm 6cm 0cm]{figures/Plots/DiffEffAcc.pdf}}
%     \caption{Accuracy for EfficientNetB3.}
%     \label{plot:Diff_Eff_Acc}
% \end{figure}

\section{Conclusion}
\label{sec:econclusion}
This paper presents \tool, a novel scheduler that enables cascade collaborative systems when assisting multiple devices at the same time. By dynamically updating the thresholds of forwarding decision functions on devices during run time, \tool manages to avoid request accumulation on server-side which keeps response latency below SLO targets and throughput high across different device pool sizes. Moreover, it considers the device heterogeneity problem and is able to maintain performance while accommodating different tiers of devices. The flexibility and scalability of \tool will help push collaborative cascade systems to enable broader DL deployment in indoor intelligent environments.

\bibliographystyle{IEEEtran}
{\footnotesize
\bibliography{references}
}

\end{document}